%% file: neurips_2024.tex
\documentclass{article}

\usepackage[nonatbib, final]{neurips_2024}

\usepackage[utf8]{inputenc} 
\usepackage[T1]{fontenc}    
\usepackage{hyperref}       
\usepackage{url}            
\usepackage{booktabs}       
\usepackage{amsfonts}       
\usepackage{nicefrac}       
\usepackage{microtype}      
\usepackage{xcolor}         
\usepackage{blindtext}
\usepackage[numbers]{natbib}
\usepackage{ circuitikz }
\usepackage{tikz}
\usepackage{graphicx}
\usepackage{tabu}
\usepackage[]{caption}

\title{
Climate Adaptation with Reinforcement Learning: Experiments with Flooding and Transportation in Copenhagen}

\author{%
    Miguel Costa, 
    Morten W. Petersen, 
    Arthur Vandervoort, \\
    \textbf{
    Martin Drews,
    Karyn Morrissey,
    Francisco C. Pereira} \\
    Department of Technology, Management and Economics\\
    Technical University of Denmark\\
    2800, Kgs. Lyngby, Denmark\\ 
    \texttt{\{migcos, s184985, apiva, mard, kamorr, camara\}@dtu.dk} \\
}

\begin{document}

\maketitle

\vspace{-4pt}
\begin{abstract}
\input{text/0.abstract}
\end{abstract}

\section{Introduction}
\label{sec:introduction}
\input{text/1.introduction}

\section{Modelling Framework}
\label{sec:framework}
\input{text/2.framework}

\section{Preliminary Experiments \& Discussion}
\label{sec:results}
\input{text/3.results}


\newpage
\begin{ack}
This work was supported by a research grant (VIL57387) from VILLUM FONDEN.
\end{ack}

\bibliographystyle{IEEEtranN}
\bibliography{references}




\end{document}

%% file: text/0.abstract.tex
\vspace{-6pt}
Due to climate change the frequency and intensity of extreme rainfall events, which contribute to urban flooding, are expected to increase in many places. These floods can damage transport infrastructure and disrupt mobility, highlighting the need for cities to adapt to escalating risks. Reinforcement learning (RL) serves as a powerful tool for uncovering optimal adaptation strategies, determining how and where to deploy adaptation measures effectively, even under significant uncertainty.
In this study, we leverage RL to identify the most effective timing and locations for implementing measures, aiming to reduce both direct and indirect impacts of flooding. Our framework integrates climate change projections of future rainfall events and floods, models city-wide motorized trips, and quantifies direct and indirect impacts on infrastructure and mobility.
Preliminary results suggest that our RL-based approach can significantly enhance decision-making by prioritizing interventions in specific urban areas and identifying the optimal periods for their implementation. 
Our framework is publicly available: \url{https://github.com/MLSM-at-DTU/floods_transport_rl}.

%% file: text/1.introduction.tex
As climate change continues to impact our world, the frequency and intensity of high impact weather events are expected to rise \citep{ipcc2023climate}. In Denmark, extreme rainfall is projected to become more severe and occur more frequently \citep{dmi2011adaptation}. As rainfall increases, so does the risk of urban pluvial flooding. Floods can significantly disrupt social and economic activities, including transportation, causing delays and loss of vehicle control \citep{wang2020climate}. To effectively address these challenges, cities must enhance their resilience.

In this work, we address the challenge of identifying the most effective adaptation measures to minimize the impacts of pluvial floods on transportation. Using Copenhagen, Denmark, as our case study, we frame our problem using a reinforcement learning (RL) approach. We build an environment that incorporates current climate projections of rainfall and consequent floods. Concurrently, we model trips which are disrupted by varying water levels, affecting mobility and transport infrastructure. To our knowledge, this is the first comprehensive framework designed to identify the best adaptation measures for enhancing transportation resilience to urban floods using such approach.

\subsection{Related Work}

Urban pluvial flooding occurs when large volumes of water accumulate in streets or roads due to insufficient drainage and infiltration capacity as a result of heavy precipitation events \citep{borowska2024changes}. Transportation is significantly impacted by such floods, both directly and indirectly \citep{lu2022overview}. These impacts include road deterioration \citep{lu2022overview}, travel delays and congestion \citep{shahdani2022assessing, ding_interregional_2023}, and loss of accessibility \citep{li2018potential, papilloud2021vulnerability}.

RL has previously been applied to a few aspects of flood management and transportation. For instance, it has been used to design emergency routing systems \citep{li2024reinforcement}, control urban drainage and stormwater systems \citep{tian2023flooding, bowes2021flood}, and study travel behaviors to inform response strategies \citep{fan2021evaluating}. However, these applications generally focus on reactive strategies -- responding to events as they occur -- rather than proactively determining the best adaptation measures to minimize future flood impacts.

On the other hand, evaluating adaptation measures typically relies on expert knowledge or limited simulations, assessing how these measures can prevent or mitigate the impacts of floods \citep{vajjarapu2020evaluating, wang2020integrated, pregnolato2016disruption, de2022climate}. To the best of our knowledge, there has been no comprehensive study using reinforcement learning of how different adaptation measures should be implemented over time to proactively minimize both the direct and indirect impacts of floods on transportation.

%% file: text/2.framework.tex
We frame our approach as an Integrated Assessment Model (IAM) that connects: 1) a rainfall projection model, 2) a flood model, 3) a transportation model, and 4) a transport infrastructure and mobility impact model. 
\autoref{fig:iam_schematic} provides an overview of our IAM framework, which we now detail.

\vspace{-12pt}
\input{figs/schematic}
\vspace{-18pt}

\subsection{Rainfall Projection Model}
Future daily rainfall statistics under the high RCP8.5 scenario \citep{vanvuuren2011} were retrieved from the Danish Meteorological Institute's Climate Atlas \citep{dmi2023klimaatlas} for the periods 2011-2040, 2041-2070, and 2071-2100. For each time slice, we assumed stationarity and formed the associated cumulative density function (CDF). Based on the CDF we sampled one heavy rainfall event per year. Urban pluvial flooding is often caused by intense precipitation of short duration (cloudbursts, from minutes to a few hours), typically associated with warm and moist conditions. For simplicity and as proof-of-concept, we assumed the projected heavy rainfall intensity (amount of precipitation) to be equal to the accumulated daily rainfall. Jointly with the choice of climate scenario, the resulting CDF is likely to overestimate the rainfall intensities and therefore represent a worst-case scenario for our periodical estimates.

\subsection{Flood Model}
After sampling a particular rainfall event (i.e., amount of rainfall), we model the associated urban flood using SCALGO Live \citep{scalgo}. SCALGO Live is a simplified interactive event-based tool for watershed delineation, and for fast modelling of flood depths and flow direction based on high-resolution digital terrain data. For Denmark, SCALGO employs the Danish Elevation Model, which is one of the world’s best national elevation models. It comprises 415 billion point data, which are used to map height differences for terrains and areas in a 0.4m grid for the entire country \citep{danishelevationmodel}. The model does not include a representation of the urban drainage system. We assumed a uniform rainfall event all over Copenhagen of unspecified duration, i.e., the water accumulates at all locations at once. Water is further distributed according to the terrain properties and filled any depressions or holes. If the volume of water exceeded the depression volume, it overflowed and continued downstream. In sum, the accumulation of heavy precipitation was mapped to water depths all around Copenhagen for identification of flooded areas.

\subsection{Transportation Model}
For the transportation component, we used a simplified version of the popular Four Step Model (4SM) \citep{mcnally2007four}, focusing exclusively on road network and motorized trips. We began by dividing Copenhagen in Traffic Analysis Zones (TAZs) following the Danish National Transport Model \citep{vejdirektoratet2022gmm}. Then, we generated and distributed trips following the distribution of trips in Copenhagen \citep{christiansen2024tu}, which were aggregated within each TAZ to map the supply and demand for each zone. In essence, this distribution reflects the underlying travel demand. Supply and demand distribution marginals were assigned used an iterative proportional fitting procedure \citep{deming1940least} with distance as a travel impedance. Lastly, we mapped routes between TAZs. TAZs’ centroids were defined as nodes and edges were created between neighboring TAZs. Paths between origin and destination TAZs were then defined as the shortest travel time paths for each trip, which can be found using Dijkstra's algorithm \citep{dijkstra1959note}. We used this network to route all trips and estimate volumes and travel times.

\subsection{Transportation and Mobility Impacts}

Finally, we computed transport and mobility impacts as three types of impacts: direct road infrastructure damage impacts and indirect impacts due to increase travel delays.

\textbf{Road infrastructure damage:} We began by downloading road network data from OpenStreetMap \citep{osm2024} using \texttt{osmnx} \citep{boeing2024modeling} from Copenhagen and computing their total construction costs per road type, number of lanes, presence of light posts and traffic lights \citet{ginkel2021flood}. Damage was then computed using depth-damage functions \citep{ginkel2021flood}, effectively mapping the percentage of damage on a road according to the water depth at its location. This damage accounts for reconstruction, repair, cleaning, and resurfacing works needed to restore roads to their original state. We aggregate direct impacts as the monetary losses at the $i$-th TAZ as $R_{i}$.

\textbf{Travel delays:} As water levels increase, travel speed is reduced, resulting in travel delays. To account for these effects we used a depth-disruption function \citep{pregnolato2017impact}, mapping decreased vehicle speeds to water depth. The speed reduction and consequent increased travel times were then modelled as economic losses using the danish travel delays value of time \citep{transportministeriet2022enhedspriser}, which we aggregated as $D_{i}$ for each TAZ.

\subsection{Reinforcement Learning}

Under the current climate uncertainty, we posit to uncover the best adaptation measures that minimize the impact of flooding events on transportation using RL. RL is a sub-field of machine learning that uses an agent-based approach to interact with an environment and achieve a certain goal \citep{sutton2018reinforcement}. Through training and by maximizing a pre-defined function (reward), the agent learns what is the best action (adaptation measure) to take. By default, the environment (as defined by the above IAM) is defined as a Markov Decision Process \citep{bellman1957markovian}, where each state is independent.

Albeit many adaptation measures can be devised, in this first work, we defined one possible measure: elevate roads by 1 meter (i.e., increase the minimum water depth needed to affect roads). At each time step, our RL agent takes an action on a TAZ and and collects information about the state of Copenhagen (e.g., precipitation event, period of time, direct and indirect impacts per TAZ, water depths on roads), effectively learning the best set of actions to take over time and space. We defined the reward function as an overall metric of economic loss, defined as:
\begin{equation}
    R = \sum_{i \in \texttt{TAZ}} 
    \beta_{R} R_{i} + 
    \beta_{D} D_{i} +
    \beta_{A} A_{i}
\end{equation}
where $R_{i}$, $D_{i}$ are as previously defined, $A_{i}$ is the cost of applying an action (i.e., cost of elevating roads by 1 meter), and weights $\beta$ ($\beta_{R}=\beta_{D}=\beta_{A}=1$) adjust for different component importances.

%% file: figs/schematic.tex
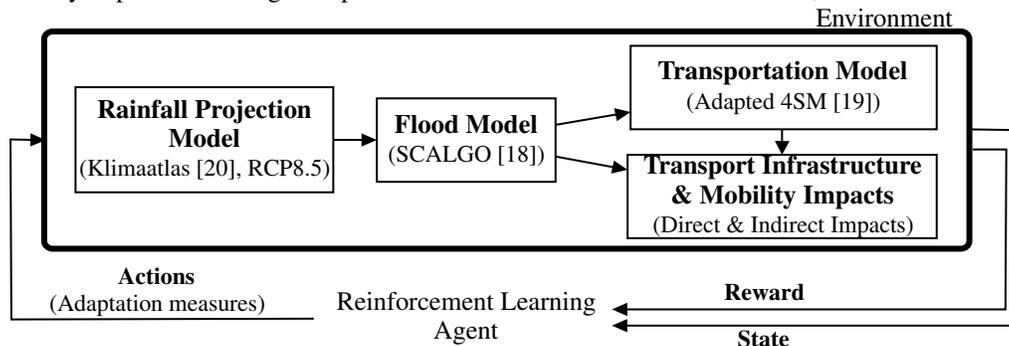
\begin{figure}[htb]
    \centering
    
\tikzset{every picture/.style={line width=0.75pt}} 

\begin{tikzpicture}[x=0.65pt,y=0.65pt,yscale=-.9,xscale=1]

\draw  [line width=2.25]  (70.6,25.4) .. controls (70.6,22.64) and (72.84,20.4) .. (75.6,20.4) -- (604,20.4) .. controls (606.76,20.4) and (609,22.64) .. (609,25.4) -- (609,156) .. controls (609,158.76) and (606.76,161) .. (604,161) -- (75.6,161) .. controls (72.84,161) and (70.6,158.76) .. (70.6,156) -- cycle ;
\draw    (611,96.75) -- (630.5,96.75) -- (630,200.25) -- (404,200.25) ;
\draw [shift={(401,200.25)}, rotate = 360] [fill={rgb, 255:red, 0; green, 0; blue, 0 }  ][line width=0.08]  [draw opacity=0] (8.93,-4.29) -- (0,0) -- (8.93,4.29) -- cycle    ;
\draw    (609,83.75) -- (639,84.25) -- (638,210.75) -- (404,210.75) ;
\draw [shift={(401,210.75)}, rotate = 360] [fill={rgb, 255:red, 0; green, 0; blue, 0 }  ][line width=0.08]  [draw opacity=0] (8.93,-4.29) -- (0,0) -- (8.93,4.29) -- cycle    ;
\draw    (227.5,206.25) -- (51.5,206.75) -- (51,90.25) -- (68.5,90.68) ;
\draw [shift={(71.5,90.75)}, rotate = 181.4] [fill={rgb, 255:red, 0; green, 0; blue, 0 }  ][line width=0.08]  [draw opacity=0] (8.93,-4.29) -- (0,0) -- (8.93,4.29) -- cycle    ;

\draw (489.25,219) node  [font=\small] [align=left] {\textbf{State}};
\draw (489.75,189) node  [font=\small] [align=left] {\textbf{Reward}};
\draw (136.25,197.5) node  [font=\small] [align=left] {\begin{minipage}[lt]{100.02pt}\setlength\topsep{0pt}
\begin{center}
{\footnotesize (Adaptation measures)}
\end{center}

\end{minipage}};
\draw (559.55,11.3) node   [align=left] {Environment};
\draw (136.25,178.9) node  [font=\small] [align=left] {\begin{minipage}[lt]{32.83pt}\setlength\topsep{0pt}
\begin{center}
\textbf{Actions}
\end{center}

\end{minipage}};
\draw    (264.2,63.7) -- (368.2,63.7) -- (368.2,117.7) -- (264.2,117.7) -- cycle  ;
\draw (316.2,90.7) node   [align=left] {\begin{minipage}[lt]{68pt}\setlength\topsep{0pt}
\begin{center}
\textbf{Flood Model}\\{\footnotesize (SCALGO \citep{scalgo})}
\end{center}

\end{minipage}};
\draw    (411.85,29.7) -- (590,29.7) -- (590,83.7) -- (411.85,83.7) -- cycle  ;
\draw (500.6,56.7) node   [align=left] {\begin{minipage}[lt]{111.04pt}\setlength\topsep{0pt}
\begin{center}
\textbf{Transportation Model}\\{\footnotesize (Adapted 4SM \citep{mcnally2007four})}
\end{center}

\end{minipage}};
\draw    (410.1,100.2) -- (590,100.2) -- (590,154.2) -- (410.1,154.2) -- cycle  ;
\draw (500.6,127.2) node   [align=left] {\begin{minipage}[lt]{105.06pt}\setlength\topsep{0pt}
\begin{center}
\textbf{Transport Infrastructure \& Mobility Impacts}\\{\footnotesize (Direct \& Indirect Impacts)}
\end{center}

\end{minipage}};
\draw    (88.85,56.47) -- (238.85,56.47) -- (238.85,124.48) -- (88.85,124.48) -- cycle  ;
\draw (163.85,90.48) node   [align=left] {\begin{minipage}[lt]{99.08pt}\setlength\topsep{0pt}
\begin{center}
\textbf{Rainfall Projection Model}\\{\footnotesize (Klimaatlas \citep{dmi2023klimaatlas}, RCP8.5)}
\end{center}

\end{minipage}};
\draw  [draw opacity=0]  (225.35,178.7) -- (407.35,178.7) -- (407.35,232.7) -- (225.35,232.7) -- cycle  ;
\draw (316.35,205.7) node   [align=left] {\begin{minipage}[lt]{120.84pt}\setlength\topsep{0pt}
\begin{center}
Reinforcement Learning Agent
\end{center}

\end{minipage}};
\draw    (368.2,80.83) -- (408.9,73.11) ;
\draw [shift={(411.85,72.55)}, rotate = 169.25] [fill={rgb, 255:red, 0; green, 0; blue, 0 }  ][line width=0.08]  [draw opacity=0] (8.93,-4.29) -- (0,0) -- (8.93,4.29) -- cycle    ;
\draw    (368.2,101.34) -- (407.16,109.31) ;
\draw [shift={(410.1,109.91)}, rotate = 191.56] [fill={rgb, 255:red, 0; green, 0; blue, 0 }  ][line width=0.08]  [draw opacity=0] (8.93,-4.29) -- (0,0) -- (8.93,4.29) -- cycle    ;
\draw    (238.85,90.59) -- (261.2,90.62) ;
\draw [shift={(264.2,90.62)}, rotate = 180.08] [fill={rgb, 255:red, 0; green, 0; blue, 0 }  ][line width=0.08]  [draw opacity=0] (8.93,-4.29) -- (0,0) -- (8.93,4.29) -- cycle    ;
\draw    (500,83.7) -- (500,97.2) ;
\draw [shift={(500,100.2)}, rotate = 270.61] [fill={rgb, 255:red, 0; green, 0; blue, 0 }  ][line width=0.08]  [draw opacity=0] (8.93,-4.29) -- (0,0) -- (8.93,4.29) -- cycle    ;

\end{tikzpicture}
    \vspace{-12pt}
    \caption{Integrated assessment model using reinforcement learning to learn what the best adaptation measures are that minimize transportation infrastructure and mobility impacts.}
    \label{fig:iam_schematic}
\end{figure}

%% file: text/3.results.tex

We setup our IAM using Python, the  Gymnasium interface \citep{towers_gymnasium_2023}, Stable-Baseline3 \citep{stable_baselines3}, and PPO \citep{huang2020closer, schulman2017proximal}. To showcase the application of our approach, we experiment with a preliminary case study. We begin by running our experiments in the city center of Copenhagen, consisting of 29 TAZ, and set the time horizon to 2023--2100. We now present preliminary results for 10 runs with distinct seeds to allow for different weather projections and increased robustness.

\input{figs/results}

\input{figs/results_per_type}

\autoref{tab:results} compares the performance of the trained RL agent against a random agent. The results demonstrate that our agent consistently outperforms the random policy, achieving significantly better outcomes overall. By 2035, our agent incurs additional costs but achieves a substantial reduction in direct and indirect impacts, lowering them by 47\%. By 2100, although both policies result in similar adaptation costs, our agent’s strategic deployment of optimal measures over time leads to significantly reduced travel delays (by 39\%) and impacts (by 45\%). \autoref{fig:results_per_impact} illustrates the impacts in 2035 for a single run. As shown, at this point, our agent has implemented specific adaptation measures in certain TAZs, resulting in lower road damages and reduced travel delays. This highlights the agent’s ability to prioritize interventions that would otherwise lead to greater losses.

These results underscore the effectiveness of using RL to identify optimal adaptation measures for Copenhagen over time, enhancing the city’s ability to address climate change more efficiently. The proposed IAM introduces a novel framework for accurately simulating future rainfall, subsequent pluvial urban floods, and their impacts on transportation and mobility. Looking ahead, we suggest that this approach could be valuable for researchers and authorities in making more efficient and informed decisions and improving urban resilience.

In future work, we plan to further develop our IAM by extending the rainfall projection model, and by expanding the case study to encompass the entire city of Copenhagen and its metropolitan area. This expansion would include additional adaptation measures (e.g., constructing permeable roads or enhancing road durability), other modes of transport (e.g., cycling and walking), broader impact categories (e.g., electric vehicle charging infrastructure, public transportation accessibility, and subjective wellbeing), and comparative analyses (e.g., expert knowledge or participatory design). Additionally, we aim to enhance our transportation simulation by more accurately modeling supply and demand dynamics and their fluctuations during flood events, which can lead to trip cancellations and increased congestion levels \citep{wang2020climate}. Including such enhancements can further refine our IAM, making it a more comprehensive and practical tool for urban resilience planning in the face of climate change.

%% file: figs/results.tex
\begin{table}[htb]
{\footnotesize
    \centering
    \caption{Differences between average reward, action costs, and impacts between random and learned policies. Two time horizons are presented: 2023--2035 coinciding with Copehagen's current climate adaptation plan \citep{copenhagen2023climateplan2035} and until 2100. Values represent the mean ± standard deviation across 10 runs.}
    \label{tab:results}
    {\tabulinesep=.5mm
    \centering
    \begin{tabu}{p{2.7cm}p{2.1cm}p{2.1cm}p{2.1cm}p{2.1cm}}
       & \multicolumn{2}{l}{2023-2035} & \multicolumn{2}{l}{2023-2100}\\ \cmidrule(l){2-3} \cmidrule(l){4-5}
       & \parbox{1.5cm}{Random Policy} & \parbox{1.5cm}{Optimal Policy} & \parbox{1.5cm}{Random Policy} & \parbox{1.5cm}{Optimal Policy}\\ \hline\hline
Reward (M DKK) $\uparrow$                        & 
\textbf{-49.19} $\pm$ 4.91 & \textbf{-36.48} $\pm$ 3.02 & \textbf{-92.09} $\pm$ 20.72 & \textbf{-69.56} $\pm$ 2.64 \\
Cumulative Cost of Measures (M DKK) $\downarrow$ & 
\textbf{ 20.08} $\pm$ 2.35 & \textbf{ 22.93} $\pm$ 1.35 & \textbf{ 52.94} $\pm$ 0.00  & \textbf{ 51.79} $\pm$ 1.47 \\
Cumulative Cost of Impacts (M DKK) $\downarrow$  & 
\textbf{ 29.10} $\pm$ 4.07 & \textbf{ 13.56} $\pm$ 2.32 & \textbf{ 39.15} $\pm$ 20.72 & \textbf{ 17.77} $\pm$ 2.12 \\
Cumulative travel delays (k h) $\downarrow$      & 
\textbf{  5.17} $\pm$ 0.79 & \textbf{  2.15} $\pm$ 0.46 & \textbf{  6.91} $\pm$ 3.98  & \textbf{  2.67} $\pm$ 0.39 \\
\hline
    \end{tabu}}
    }
\vspace{-6pt}
\end{table}

%% file: figs/results_per_type.tex
\begin{figure}[htb]
  \centering
  \begin{minipage}{.25\textwidth}
    \centering
    \includegraphics[trim={0cm 0cm 0cm 0cm},clip, width=.98\textwidth]{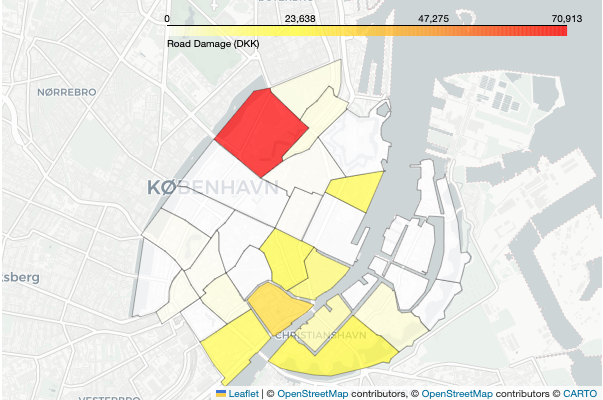} 
  \end{minipage}%
  \begin{minipage}{.25\textwidth}
    \centering
    \includegraphics[trim={0cm 0cm 0cm 0cm},clip, width=.98\textwidth]{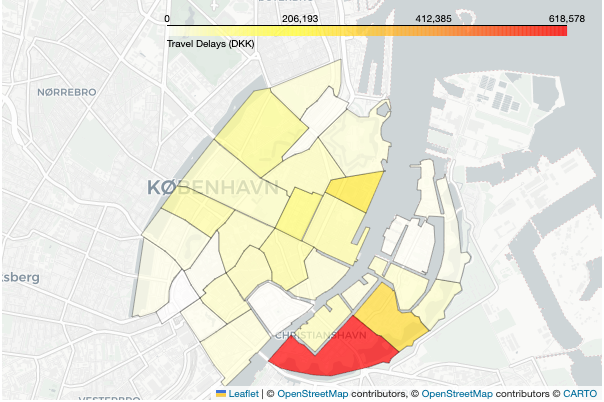} 
  \end{minipage}%
  \begin{minipage}{.25\textwidth}
    \centering
    \includegraphics[trim={0cm 0cm 0cm 0cm},clip, width=.98\textwidth]{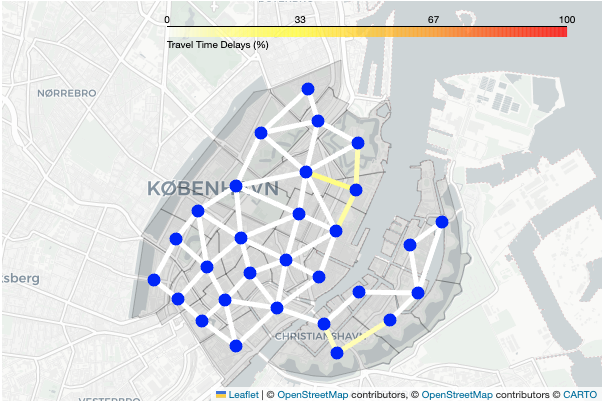} 
  \end{minipage}%
  \begin{minipage}{.25\textwidth}
    \centering
    \includegraphics[trim={0cm 0cm 0cm 0cm},clip, width=.98\textwidth]{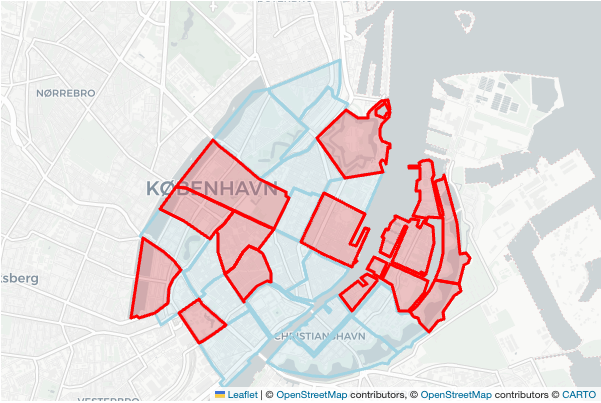} 
  \end{minipage}%

  \begin{minipage}{.25\textwidth}
    \centering
    \includegraphics[trim={0cm 0cm 0cm 0cm},clip, width=.98\textwidth]{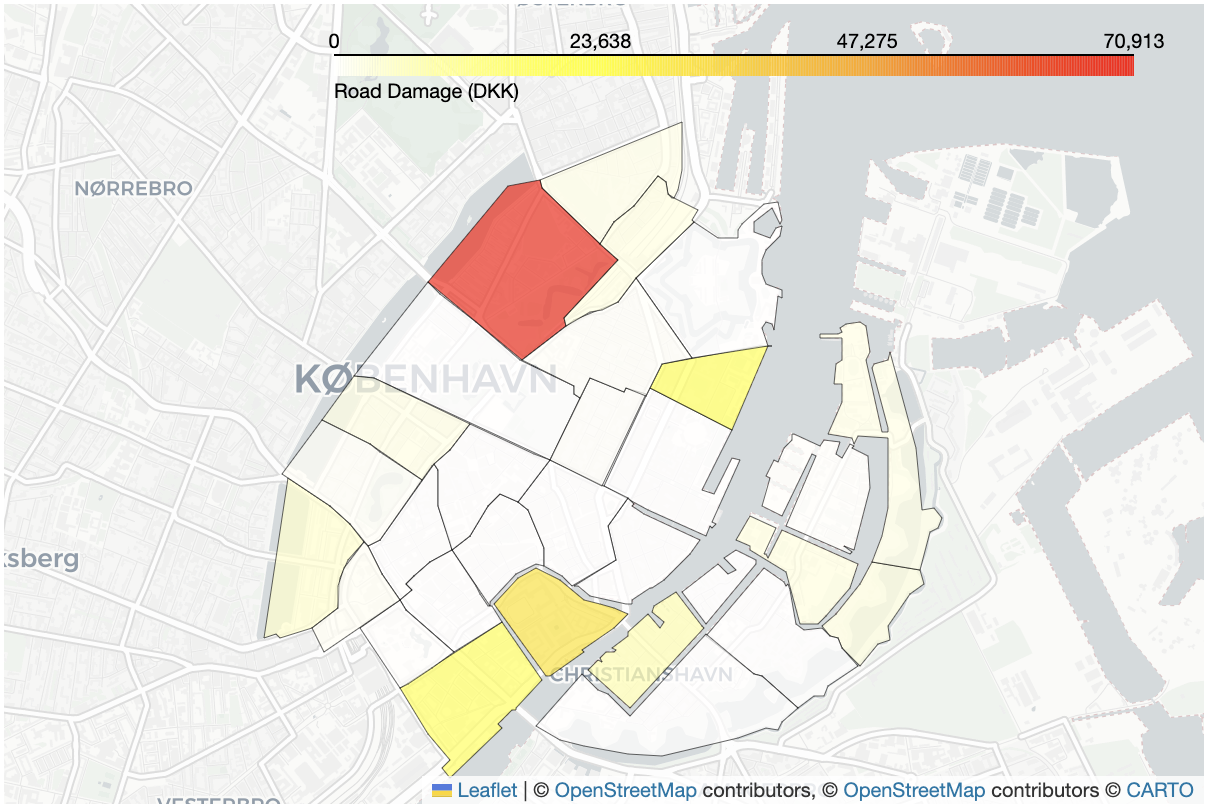} 
  \end{minipage}%
  \begin{minipage}{.25\textwidth}
    \centering
    \includegraphics[trim={0cm 0cm 0cm 0cm},clip, width=.98\textwidth]{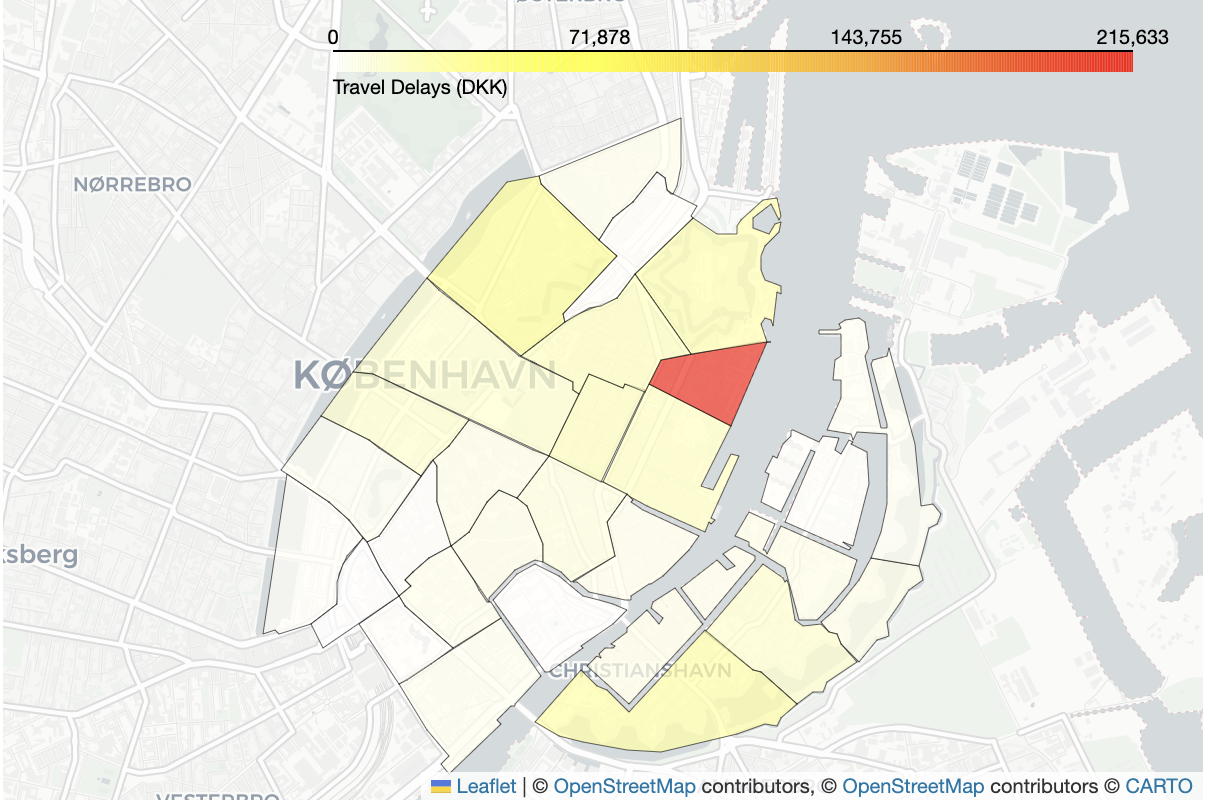} 
  \end{minipage}%
  \begin{minipage}{.25\textwidth}
    \centering
    \includegraphics[trim={0cm 0cm 0cm 0cm},clip, width=.98\textwidth]{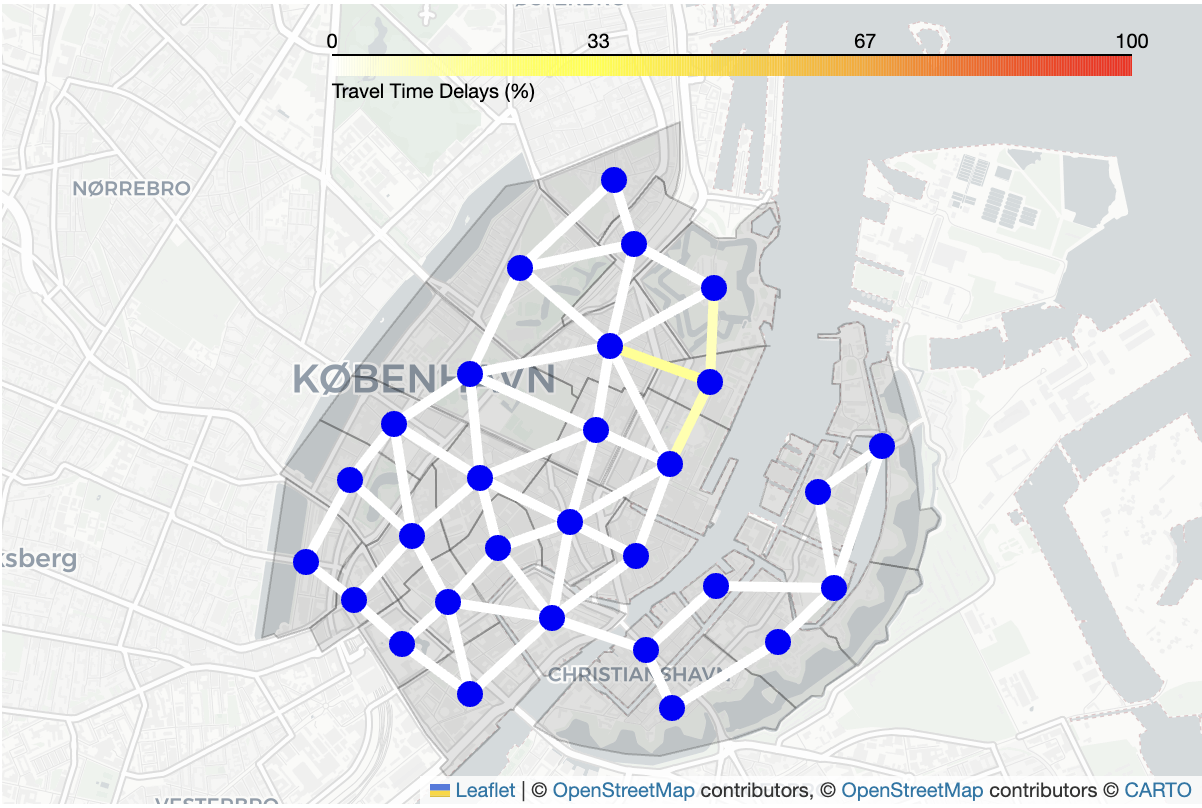} 
  \end{minipage}%
  \begin{minipage}{.25\textwidth}
    \centering
    \includegraphics[trim={0cm 0cm 0cm 0cm},clip, width=.98\textwidth]{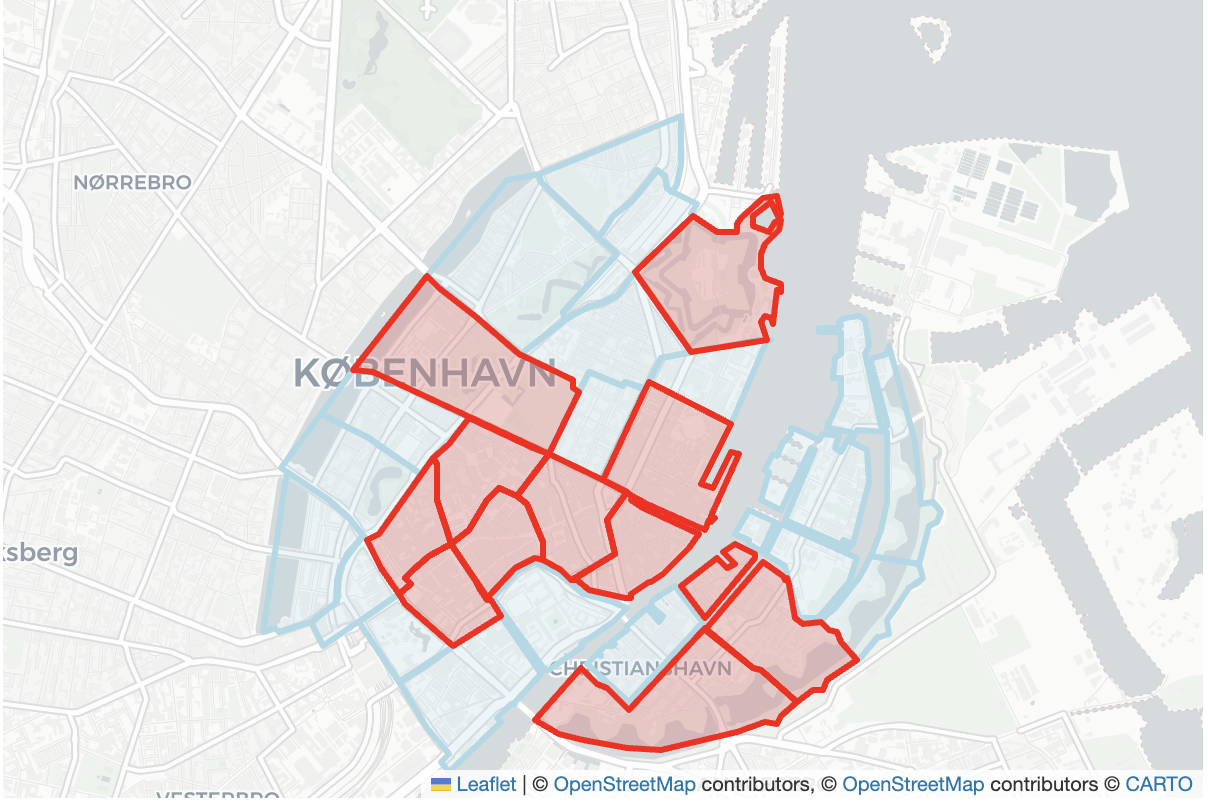} 
  \end{minipage}%
  
\caption{Costs of floods impacts on transportation and mobility in Copenhagen in 2035. Top row shows results with random adaptation measures deployed over time and space, while bottom row shows impacts using optimal adaptations over time. From left to right: direct road infrastructure impacts, indirect impacts as travel delays, percentage of travel time difference for travel between TAZ, and where adaptation measures were deployed (red).}
\label{fig:results_per_impact}
\vspace{-10pt}
\end{figure}

%% file: neurips_2024.bbl
\begin{thebibliography}{38}
\providecommand{\natexlab}[1]{#1}
\providecommand{\url}[1]{#1}
\csname url@samestyle\endcsname
\providecommand{\newblock}{\relax}
\providecommand{\bibinfo}[2]{#2}
\providecommand{\BIBentrySTDinterwordspacing}{\spaceskip=0pt\relax}
\providecommand{\BIBentryALTinterwordstretchfactor}{4}
\providecommand{\BIBentryALTinterwordspacing}{\spaceskip=\fontdimen2\font plus
\BIBentryALTinterwordstretchfactor\fontdimen3\font minus
  \fontdimen4\font\relax}
\providecommand{\BIBforeignlanguage}[2]{{%
\expandafter\ifx\csname l@#1\endcsname\relax
\typeout{** WARNING: IEEEtranN.bst: No hyphenation pattern has been}%
\typeout{** loaded for the language `#1'. Using the pattern for}%
\typeout{** the default language instead.}%
\else
\language=\csname l@#1\endcsname
\fi
#2}}
\providecommand{\BIBdecl}{\relax}
\BIBdecl

\bibitem[{IPCC}(2023)]{ipcc2023climate}
{IPCC}, ``{Section 3: Long-Term Climate and Development Futures},'' in
  \emph{Climate Change 2023: Synthesis Report. Contribution of Working Groups
  I, II and III to the Sixth Assessment Report of the Intergovernmental Panel
  on Climate Change [Core Writing Team, H. Lee and J. Romero (eds.)]}.\hskip
  1em plus 0.5em minus 0.4em\relax IPCC, Geneva, Switzerland, doi:
  10.59327/IPCC/AR6-9789291691647, 2023, pp. 35--115.

\bibitem[{Danmarks Meteorologiske Institut}(2011)]{dmi2011adaptation}
{Danmarks Meteorologiske Institut}, ``{Adaptation to the future climate in
  Denmark},''
  \url{https://en.klimatilpasning.dk/media/7863/klimatilpasningshæfte%20uk%20web.pdf},
  2011, accessed: 2024-06-14.

\bibitem[Wang et~al.(2020{\natexlab{a}})Wang, Qu, Yang, Nichol, Clarke, and
  Ge]{wang2020climate}
T.~Wang, Z.~Qu, Z.~Yang, T.~Nichol, G.~Clarke, and Y.-E. Ge, ``Climate change
  research on transportation systems: Climate risks, adaptation and planning,''
  \emph{Transportation Research Part D: Transport and Environment}, vol.~88, p.
  102553, 2020.

\bibitem[Borowska-Stefańska et~al.(2024)Borowska-Stefańska, Bartnik,
  Dulebenets, Kowalski, Sahebgharani, Tomalski, and
  Wiśniewski]{borowska2024changes}
\BIBentryALTinterwordspacing
M.~Borowska-Stefańska, A.~Bartnik, M.~A. Dulebenets, M.~Kowalski,
  A.~Sahebgharani, P.~Tomalski, and S.~Wiśniewski, ``Changes in intra-city
  transport accessibility accompanying the occurrence of an urban flood,''
  \emph{Transportation Research Part D: Transport and Environment}, vol. 126,
  p. 104040, 2024. [Online]. Available:
  \url{https://www.sciencedirect.com/science/article/pii/S1361920923004376}
\BIBentrySTDinterwordspacing

\bibitem[Lu et~al.(2022)Lu, {Shun Chan}, Chen, Chan, and Gu]{lu2022overview}
\BIBentryALTinterwordspacing
X.~Lu, F.~K. {Shun Chan}, W.-Q. Chen, H.~K. Chan, and X.~Gu, ``An overview of
  flood-induced transport disruptions on urban streets and roads in chinese
  megacities: Lessons and future agendas,'' \emph{Journal of Environmental
  Management}, vol. 321, p. 115991, 2022. [Online]. Available:
  \url{https://www.sciencedirect.com/science/article/pii/S030147972201564X}
\BIBentrySTDinterwordspacing

\bibitem[Shahdani et~al.(2022)Shahdani, Santamaria-Ariza, Sousa, Coelho, and
  Matos]{shahdani2022assessing}
\BIBentryALTinterwordspacing
F.~J. Shahdani, M.~Santamaria-Ariza, H.~S. Sousa, M.~Coelho, and J.~C. Matos,
  ``{Assessing Flood Indirect Impacts on Road Transport Networks Applying
  Mesoscopic Traffic Modelling: The Case Study of Santarém, Portugal},''
  \emph{Applied Sciences}, vol.~12, no.~6, 2022. [Online]. Available:
  \url{https://www.mdpi.com/2076-3417/12/6/3076}
\BIBentrySTDinterwordspacing

\bibitem[Ding and Wu(2023)]{ding_interregional_2023}
\BIBentryALTinterwordspacing
W.~Ding and J.~Wu, ``\BIBforeignlanguage{en}{Interregional economic impacts of
  an extreme storm flood scenario considering transportation interruption: {A}
  case study of {Shanghai}, {China}},''
  \emph{\BIBforeignlanguage{en}{Sustainable Cities and Society}}, vol.~88, p.
  104296, Jan. 2023. [Online]. Available:
  \url{https://linkinghub.elsevier.com/retrieve/pii/S221067072200600X}
\BIBentrySTDinterwordspacing

\bibitem[Li et~al.(2018)Li, Kwan, Yin, Yu, and Wang]{li2018potential}
\BIBentryALTinterwordspacing
M.~Li, M.-P. Kwan, J.~Yin, D.~Yu, and J.~Wang, ``{The potential effect of a
  100-year pluvial flood event on metro accessibility and ridership: A case
  study of central Shanghai, China},'' \emph{Applied Geography}, vol. 100, pp.
  21--29, 2018. [Online]. Available:
  \url{https://www.sciencedirect.com/science/article/pii/S0143622818302716}
\BIBentrySTDinterwordspacing

\bibitem[Papilloud and Keiler(2021)]{papilloud2021vulnerability}
\BIBentryALTinterwordspacing
T.~Papilloud and M.~Keiler, ``Vulnerability patterns of road network to extreme
  floods based on accessibility measures,'' \emph{Transportation Research Part
  D: Transport and Environment}, vol. 100, p. 103045, 2021. [Online].
  Available:
  \url{https://www.sciencedirect.com/science/article/pii/S1361920921003424}
\BIBentrySTDinterwordspacing

\bibitem[Li et~al.(2024)Li, Zhang, Alizadeh, Zhang, Duffield, Meyer, Thompson,
  Gao, and Behzadan]{li2024reinforcement}
D.~Li, Z.~Zhang, B.~Alizadeh, Z.~Zhang, N.~Duffield, M.~A. Meyer, C.~M.
  Thompson, H.~Gao, and A.~H. Behzadan, ``A reinforcement learning-based
  routing algorithm for large street networks,'' \emph{International Journal of
  Geographical Information Science}, vol.~38, no.~2, pp. 183--215, 2024.

\bibitem[Tian et~al.(2023)Tian, Xin, Zhang, Zhao, Liao, and
  Tao]{tian2023flooding}
W.~Tian, K.~Xin, Z.~Zhang, M.~Zhao, Z.~Liao, and T.~Tao, ``Flooding mitigation
  through safe \& trustworthy reinforcement learning,'' \emph{Journal of
  Hydrology}, vol. 620, p. 129435, 2023.

\bibitem[Bowes et~al.(2021)Bowes, Tavakoli, Wang, Heydarian, Behl, Beling, and
  Goodall]{bowes2021flood}
B.~D. Bowes, A.~Tavakoli, C.~Wang, A.~Heydarian, M.~Behl, P.~A. Beling, and
  J.~L. Goodall, ``Flood mitigation in coastal urban catchments using real-time
  stormwater infrastructure control and reinforcement learning,'' \emph{Journal
  of Hydroinformatics}, vol.~23, no.~3, pp. 529--547, 2021.

\bibitem[Fan et~al.(2021)Fan, Jiang, and Mostafavi]{fan2021evaluating}
C.~Fan, X.~Jiang, and A.~Mostafavi, ``Evaluating crisis perturbations on urban
  mobility using adaptive reinforcement learning,'' \emph{Sustainable Cities
  and Society}, vol.~75, p. 103367, 2021.

\bibitem[Vajjarapu et~al.(2020)Vajjarapu, Verma, and
  Allirani]{vajjarapu2020evaluating}
H.~Vajjarapu, A.~Verma, and H.~Allirani, ``Evaluating climate change adaptation
  policies for urban transportation in india,'' \emph{International journal of
  disaster risk reduction}, vol.~47, p. 101528, 2020.

\bibitem[Wang et~al.(2020{\natexlab{b}})Wang, Yang, Gao, Hu, Zhao, and
  Stanley]{wang2020integrated}
W.~Wang, S.~Yang, J.~Gao, F.~Hu, W.~Zhao, and H.~E. Stanley, ``{An Integrated
  Approach for Assessing the Impact of Large-Scale Future Floods on a Highway
  Transport System},'' \emph{Risk analysis}, vol.~40, no.~9, pp. 1780--1794,
  2020.

\bibitem[Pregnolato et~al.(2016)Pregnolato, Ford, and
  Dawson]{pregnolato2016disruption}
M.~Pregnolato, A.~Ford, and R.~Dawson, ``Disruption and adaptation of urban
  transport networks from flooding,'' in \emph{E3s Web of conferences},
  vol.~7.\hskip 1em plus 0.5em minus 0.4em\relax EDP Sciences, 2016, p. 07006.

\bibitem[de~Abreu et~al.(2022)de~Abreu, Santos, and Monteiro]{de2022climate}
V.~H.~S. de~Abreu, A.~S. Santos, and T.~G.~M. Monteiro, ``Climate change
  impacts on the road transport infrastructure: A systematic review on
  adaptation measures,'' \emph{Sustainability}, vol.~14, no.~14, p. 8864, 2022.

\bibitem[SCALGO(2024)]{scalgo}
\BIBentryALTinterwordspacing
SCALGO, ``{SCALGO Live},'' 2024, {Accessed: 2024-06-07}. [Online]. Available:
  \url{https://scalgo.com/live/denmark}
\BIBentrySTDinterwordspacing

\bibitem[McNally(2007)]{mcnally2007four}
M.~G. McNally, ``The four-step model,'' in \emph{Handbook of transport
  modelling}.\hskip 1em plus 0.5em minus 0.4em\relax Emerald Group Publishing
  Limited, 2007, vol.~1, pp. 35--53.

\bibitem[{Danmarks Meteorologiske Institut}(2023)]{dmi2023klimaatlas}
\BIBentryALTinterwordspacing
{Danmarks Meteorologiske Institut}, ``Klimaatlas,'' 2023, {Accessed:
  2024-08-26}. [Online]. Available:
  \url{https://www.dmi.dk/klima-atlas/data-i-klimaatlas}
\BIBentrySTDinterwordspacing

\bibitem[Van~Vuuren et~al.(2011)Van~Vuuren, Edmonds, Kainuma, Riahi, Thomson,
  Hibbard, Hurtt, Kram, Krey, Lamarque, et~al.]{vanvuuren2011}
D.~P. Van~Vuuren, J.~Edmonds, M.~Kainuma, K.~Riahi, A.~Thomson, K.~Hibbard,
  G.~C. Hurtt, T.~Kram, V.~Krey, J.-F. Lamarque \emph{et~al.}, ``The
  representative concentration pathways: an overview,'' \emph{Climatic change},
  vol. 109, pp. 5--31, 2011.

\bibitem[{Styrelsen for Dataforsyning og
  Infrastruktur}(2024)]{danishelevationmodel}
\BIBentryALTinterwordspacing
{Styrelsen for Dataforsyning og Infrastruktur}, ``{Danmarks Højdemodel},''
  2024, accessed: 2024-06-07. [Online]. Available:
  \url{https://sdfi.dk/data-om-danmark/vores-data/danmarks-hoejdemodel}
\BIBentrySTDinterwordspacing

\bibitem[Vejdirektoratet(2022)]{vejdirektoratet2022gmm}
\BIBentryALTinterwordspacing
Vejdirektoratet, ``{Grøn Mobilitetsmodel (GMM)},'' 2022, {Accessed:
  2024-06-06}. [Online]. Available:
  \url{https://www.vejdirektoratet.dk/segment/groen-mobilitetsmodel}
\BIBentrySTDinterwordspacing

\bibitem[Christiansen and Baescu(2024)]{christiansen2024tu}
H.~Christiansen and O.~Baescu, ``The danish national travel survey: 0623v1,''
  \url{https://doi.org/10.11581/dtu:00000034}, 2024, {Accessed: 2024-06-01}.

\bibitem[Deming and Stephan(1940)]{deming1940least}
W.~E. Deming and F.~F. Stephan, ``On a least squares adjustment of a sampled
  frequency table when the expected marginal totals are known,'' \emph{The
  Annals of Mathematical Statistics}, vol.~11, no.~4, pp. 427--444, 1940.

\bibitem[Dijkstra(1959)]{dijkstra1959note}
E.~W. Dijkstra, ``A note on two problems in connexion with graphs,''
  \emph{Numerische mathematik}, vol.~1, no.~1, pp. 269--271, 1959.

\bibitem[{OpenStreetMap contributors}(2024)]{osm2024}
{OpenStreetMap contributors}, ``Openstreetmap foundation. available as open
  data under the open data commons open database license (odbl) at
  \url{openstreetmap.org},'' 2024.

\bibitem[Boeing(2024)]{boeing2024modeling}
G.~Boeing, ``{Modeling and Analyzing Urban Networks and Amenities with
  OSMnx},'' 2024.

\bibitem[van Ginkel et~al.(2021)van Ginkel, Dottori, Alfieri, Feyen, and
  Koks]{ginkel2021flood}
\BIBentryALTinterwordspacing
K.~C.~H. van Ginkel, F.~Dottori, L.~Alfieri, L.~Feyen, and E.~E. Koks, ``Flood
  risk assessment of the european road network,'' \emph{Natural Hazards and
  Earth System Sciences}, vol.~21, no.~3, pp. 1011--1027, 2021. [Online].
  Available: \url{https://nhess.copernicus.org/articles/21/1011/2021/}
\BIBentrySTDinterwordspacing

\bibitem[Pregnolato et~al.(2017)Pregnolato, Ford, Wilkinson, and
  Dawson]{pregnolato2017impact}
\BIBentryALTinterwordspacing
M.~Pregnolato, A.~Ford, S.~M. Wilkinson, and R.~J. Dawson, ``The impact of
  flooding on road transport: A depth-disruption function,''
  \emph{Transportation Research Part D: Transport and Environment}, vol.~55,
  pp. 67--81, 2017. [Online]. Available:
  \url{https://www.sciencedirect.com/science/article/pii/S1361920916308367}
\BIBentrySTDinterwordspacing

\bibitem[Transportministeriet(2022)]{transportministeriet2022enhedspriser}
Transportministeriet, ``Transportøkonomiske enhedspriser, version 2.0,''
  \url{https://www.man.dtu.dk/myndighedsbetjening/teresa-og-transportoekonomiske-enhedspriser},
  2022.

\bibitem[Sutton and Barto(2018)]{sutton2018reinforcement}
R.~S. Sutton and A.~G. Barto, \emph{Reinforcement learning: An
  introduction}.\hskip 1em plus 0.5em minus 0.4em\relax MIT press, 2018.

\bibitem[Bellman(1957)]{bellman1957markovian}
R.~Bellman, ``A markovian decision process,'' \emph{Journal of Mathematics and
  Mechanics}, pp. 679--684, 1957.

\bibitem[Towers et~al.(2023)Towers, Terry, Kwiatkowski, Balis, Cola, Deleu,
  Goulão, Kallinteris, KG, Krimmel, Perez-Vicente, Pierré, Schulhoff, Tai,
  Shen, and Younis]{towers_gymnasium_2023}
\BIBentryALTinterwordspacing
M.~Towers, J.~K. Terry, A.~Kwiatkowski, J.~U. Balis, G.~d. Cola, T.~Deleu,
  M.~Goulão, A.~Kallinteris, A.~KG, M.~Krimmel, R.~Perez-Vicente, A.~Pierré,
  S.~Schulhoff, J.~J. Tai, A.~T.~J. Shen, and O.~G. Younis, ``Gymnasium,'' Mar.
  2023. [Online]. Available: \url{https://zenodo.org/record/8127025}
\BIBentrySTDinterwordspacing

\bibitem[Raffin et~al.(2021)Raffin, Hill, Gleave, Kanervisto, Ernestus, and
  Dormann]{stable_baselines3}
\BIBentryALTinterwordspacing
A.~Raffin, A.~Hill, A.~Gleave, A.~Kanervisto, M.~Ernestus, and N.~Dormann,
  ``{Stable-Baselines3: Reliable Reinforcement Learning Implementations},''
  \emph{Journal of Machine Learning Research}, vol.~22, no. 268, pp. 1--8,
  2021. [Online]. Available: \url{http://jmlr.org/papers/v22/20-1364.html}
\BIBentrySTDinterwordspacing

\bibitem[Huang and Onta{\~n}{\'o}n(2020)]{huang2020closer}
S.~Huang and S.~Onta{\~n}{\'o}n, ``A closer look at invalid action masking in
  policy gradient algorithms,'' \emph{arXiv preprint arXiv:2006.14171}, 2020.

\bibitem[Schulman et~al.(2017)Schulman, Wolski, Dhariwal, Radford, and
  Klimov]{schulman2017proximal}
J.~Schulman, F.~Wolski, P.~Dhariwal, A.~Radford, and O.~Klimov, ``Proximal
  policy optimization algorithms,'' \emph{arXiv preprint arXiv:1707.06347},
  2017.

\bibitem[{City of Copenhagen}(2023)]{copenhagen2023climateplan2035}
\BIBentryALTinterwordspacing
{City of Copenhagen}, ``Climate plan 2035,'' 2023, {Accessed: 2024-08-27}.
  [Online]. Available:
  \url{https://international.kk.dk/about-copenhagen/liveable-green-city/2035-climate-plan}
\BIBentrySTDinterwordspacing

\end{thebibliography}
